\documentclass[10pt,twocolumn,letterpaper]{article}

\usepackage{cvpr}
\usepackage{times}
\usepackage{epsfig}
\usepackage{graphicx}
\usepackage{amsmath}
\usepackage{amssymb}
\usepackage{booktabs}
\usepackage{subfig}
\usepackage{paralist}
\usepackage{animate}
\usepackage{titling}

\usepackage[pagebackref=true,breaklinks=true,letterpaper=true,colorlinks,bookmarks=false,citecolor=blue,linkcolor=blue]{hyperref}
\cvprfinalcopy 

\def\cvprPaperID{4906} 

\date{\vspace{-5ex}}
\ifcvprfinal\fi
\begin{document}

\title{Inserting Videos into Videos}

\author{
  Donghoon Lee$^{1,2}$ \qquad
  Tomas Pfister$^2$ \qquad
  Ming-Hsuan Yang$^{2,3}$ \\
  $^1$Electrical and Computer Engineering and ASRI, Seoul National University\\
  $^2$Google Cloud AI \\
  $^3$Electrical Engineering and Computer Science, University of California at Merced
}

\maketitle

\begin{figure*}[h!]
    \centering
    \includegraphics[width=.93\linewidth]{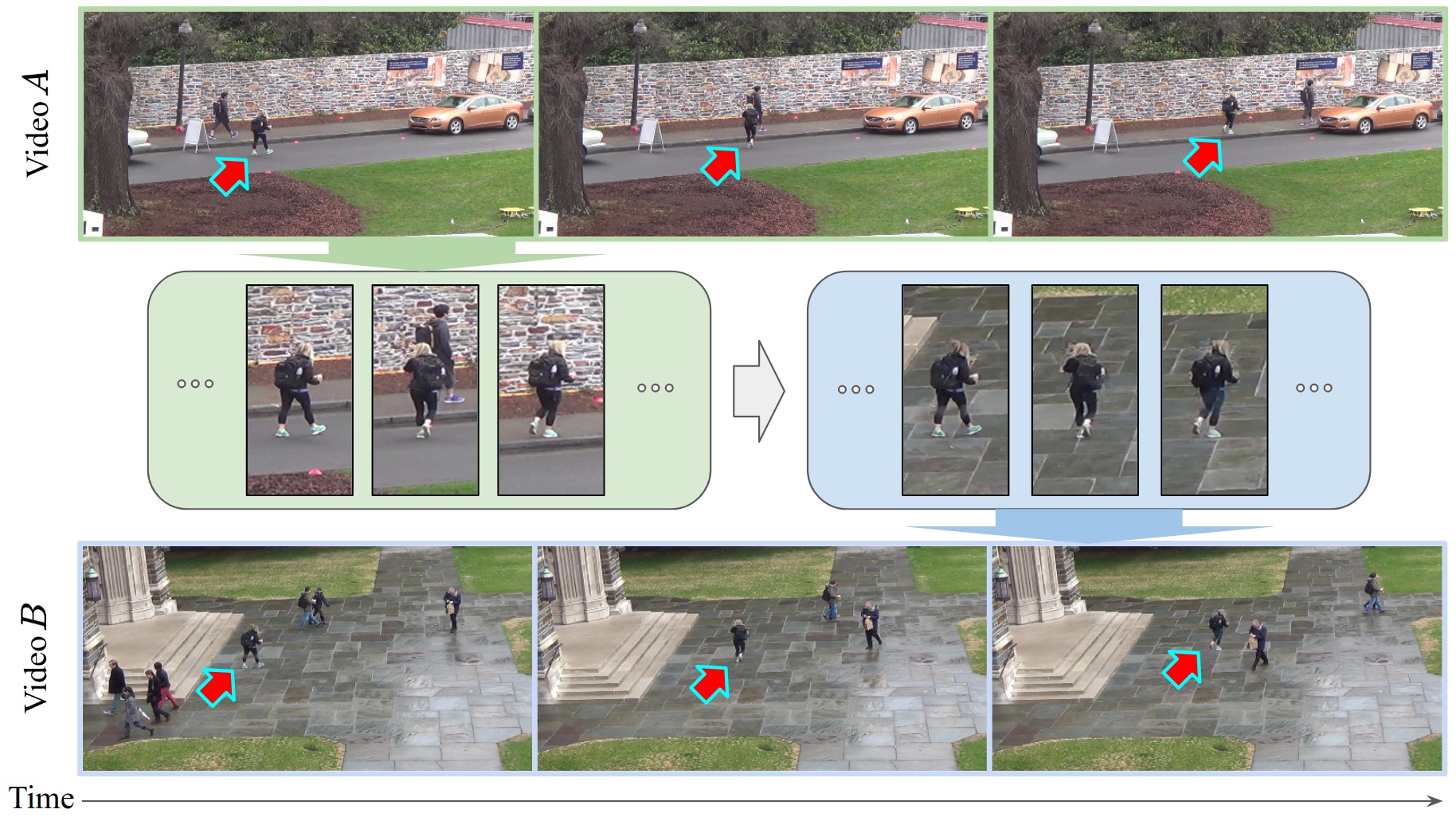}
    \caption{
    Given two videos, our algorithm aims to insert an object from one video to the other video.
    Red arrows point the inserted object originally exists in video $A$.
    We cast the problem as a video-to-video synthesis task.
    The proposed algorithm does not rely on any external domain knowledge such as the semantic segmentation mask or body pose.
    }
    \vspace{-4mm}
\label{fig:teaser}
\end{figure*}

\begin{abstract}
In this paper, we introduce a new problem of manipulating a given video by inserting other videos into it.
Our main task is, given an object video and a scene video, to insert the object video at a user-specified location in the scene video so that the resulting video looks realistic.
We aim to handle different object motions and complex backgrounds without expensive segmentation annotations.
As it is difficult to collect training pairs for this problem, we synthesize fake training pairs that can provide helpful supervisory signals when training a neural network with unpaired real data.
The proposed network architecture can take both real and fake pairs as input and perform both supervised and unsupervised training in an adversarial learning scheme.
To synthesize a realistic video, the network renders each frame based on the current input and previous frames.
Within this framework, we observe that injecting noise into previous frames while generating the current frame stabilizes training.
We conduct experiments on real-world videos in object tracking and person re-identification benchmark datasets.
Experimental results demonstrate that the proposed algorithm is able to synthesize long sequences of realistic videos with a given object video inserted.
\end{abstract}

\section{Introduction}
Object insertion in images aims to insert a new object into a given scene such that the manipulated scene looks realistic.
In recent years, there has been increasing interest in this problem as it can be applied to numerous vision tasks, including but not limited to training data augmentation for object detection~\cite{ouyang2018pedestrian}, interactive image editing~\cite{hong2018learning}, and manipulating semantic layouts~\cite{lee2018context}.
However, there remains a significant gap between its potential and real-world applications since existing methods focus on modifying a single image while either requiring carefully pre-processed inputs, \eg segmented objects without backgrounds~\cite{lin2018stgan}, or generating objects from a random vector which makes it difficult to control the resulting appearance of the object directly~\cite{hong2018learning,lee2018context,ouyang2018pedestrian}.

In this paper, we introduce a new problem of inserting existing videos into other videos.
More specifically, as shown in Figure~\ref{fig:teaser}, a user can select a video of an object of interest, \eg a walking pedestrian, and put it at a desired location in other videos, \eg surveillance scenes.
Then, an algorithm composes the object seamlessly while it moves in the scene video.
Note that unlike previous approaches \cite{hong2018learning,lee2018context,ouyang2018pedestrian}, we do not assume that the input videos have expensive segmentation annotations.
This not only allows users to edit videos more directly and intuitively, but also opens the door to numerous applications from training data augmentation for object tracking, video person re-identification, and video object segmentation, to video content generation for virtual reality or movies.

We pose the problem as a video-to-video synthesis task where the synthesized video containing an object of interest should follow the distribution of existing objects in the scene video.
This falls into an unsupervised video-to-video translation problem since we do not have paired data in general, \ie we do not observe exactly the same motion of the same object at the location we want to insert in different videos.
Nevertheless, without any supervision, we face challenging issues such as handling different backgrounds, occlusions, lighting conditions and object sizes.
Existing methods are limited to addressing such issues when there exists a number of moving objects and complex backgrounds.
For example, the performance of an algorithm that relies on object segmentation methods, which often fails to crop foreground objects accurately in a complex scene, will be bounded by the accuracy of the segmentation algorithm.

To address the problem, we first address the related problems in the image domain, \ie we study how to insert a given object image into other frames from different videos.
To alleviate the issue of unpaired data, we propose a simple yet effective way to synthesize fake data that can provide supervisory signals for object insertion.
The key idea of this supervision approach using the fake data is, when training a network, the fake data is carefully rendered to closely match the distribution of real data so that back-propagated gradient signals from the supervised fake data can help training the network with the unsupervised real data.
In this work, the fake data is generated by blending an object image and a random background patch from each video.
Then, the network learns how to reconstruct the object from the blended data.
As the reconstruction errors provide strong supervisory signals, this approach facilitates  the learning process of the generative adversarial framework \cite{goodfellow2014generative} using unpaired real data.
During inference, a new object is blended into a target location of the scene video and then fed to the trained network.

To extend the above-described algorithm to videos, we discuss how to utilize a history of synthesized frames to obtain a temporally consistent video.
We observe that if we simply add a history of previous frames as a new source of input to the object insertion network trained on images, the network will easily collapse by relying only on the (clean) previous frames instead of the (blended) current frame.
To avoid this pitfall, we use an idea from the denoising autoencoder~\cite{vincent2008extracting}: a random noise is injected into previous frames before synthesizing the current frame.
It forces the network to learn semantics between previous frames and the current input instead of blindly copy-and-pasting most of the information from the previous frames.

We conduct extensive experiments with strong baseline methods to evaluate the effectiveness of the proposed algorithm on real-world data.
Experimental results show that the proposed algorithm can insert challenging objects, \eg moving pedestrians under the cluttered backgrounds, into other videos.
For quantitative evaluation, we carry out three experiments.
First, we measure the recall of the state-of-the-art object detector \cite{redmon2018yolov3} for the inserted object.
It assesses the overall appearance of the inserted object given the surrounding context.
Second, given the state-of-the-art segmentation algorithm \cite{deeplabv3plus2018}, we measure pixel-level precision and recall of the inserted object.
Third, we perform a human subjective study for evaluating the realism of inserted objects.

The main contributions of this work are summarized as follows:
\begin{compactitem}
    \item We introduce an important and challenging problem which broadens the domain of object insertion from images to videos.
    \item We propose a novel approach to synthesize supervised fake training pairs that can help a deep neural network to learn to insert objects without supervised real pairs.
    \item We develop a new conditional GAN model to facilitate the joint training of both unsupervised real and supervised fake training pairs.
    \item We demonstrate that the proposed algorithm can synthesize realistic videos based on challenging real-world input videos.
\end{compactitem}

\section{Related Work}
\paragraph{Inserting objects into images.}
Given a pair of an object image and a scene image, the ST-GAN approach~\cite{lin2018stgan} learns a warping of the object conditioned on the scene.
Based on the warping, the object is transformed to a new location without changing its appearance.
As it focuses on geometric realism, they use carefully segmented object as an input.

Other approaches aim to insert an object by rendering its appearance.
In \cite{hong2018learning}, an object in a target category is inserted into a scene given a location and a size of a bounding box. 
It first predicts a shape of the object in the semantic space, after which an output image is generated from the predicted semantic label map and an input image.
A similar approach is proposed in~\cite{ouyang2018pedestrian} without using a semantic label map.
A bounding box of a pedestrian is replaced by random noise and then infilled with a new pedestrian based on the surrounding context.

To learn both placement and shape of a new object, the method in~\cite{chien2017detecting} removes existing objects from the scene using an image in-painting algorithm.
Then, a network is trained to recover the existing objects.
The results of this method rely significantly run script on whether the adopted image in-painting algorithm performs well, \eg not generating noisy pixels.
This issue is alleviated in \cite{lee2018context} by learning the joint distribution of the location and shape of an object conditioned on the semantic label map.
This method aims to find plausible locations and sizes of a bounding box by learning diverse affine transforms that warp a unit bounding box into the scene.
Then, objects of different shapes are synthesized conditioned on the predicted location and its surrounding context.

In contrast to existing methods, our algorithm allows a user to specify both the appearance of an object to insert and its location.
In addition, our algorithm does not require a segmentation map for training or test.

\vspace{-4mm}
\paragraph{Conditional video synthesis.}
The future frame prediction task conditions on previous frames to synthesize image content~\cite{mathieu2015deep,finn2016unsupervised,walker2016uncertain,denton2017unsupervised,liang2017dual,villegas2017decomposing,villegas2017learning}.
Due to the future uncertainty and accumulated error in the prediction process, it typically can generate only short video sequences.
On the other hand, we synthesize long video sequences by inserting one video into other videos.

The contents of a video can be transferred to other videos to synthesize new videos.
In~\cite{chan2018everybody}, given a source video of a person, the method transfers one's motion to another person in the target video.
This method estimates the object motion using a detected body pose and trains a network to render a person conditioned on the pose.
The trained network renders a new video as if the target subject follows the motion of the source video.
Instead of following exactly the same motion, the approach in \cite{bansal2018recycle} transfers an abstract content of the source video while the style of the target video is preserved.
A cyclic spatio-temporal constraint is proposed to address the task in an unsupervised manner.
It translates a source frame to a target domain and predicts the next frame.
Then, the predicted frame is translated back to the source domain.
This work also forms a cyclic loop which can improve the video quality.

The dynamic contents/textures in a video can also be used for conditional video synthesis.
In \cite{tesfaldet2018}, dynamic textures in a video such as water flow or fire flame are captured by learning a two-stream network.
Then, the work animates an input image to a video with realistic dynamic motions.
Artistic styles of a video is transferred to edit a target video while preserving its contents~\cite{huang2017real,ruder2018artistic}.

For more generic video-to-video translations, the scheme in~\cite{wang2018video} formulates conditional generative adversarial networks (GANs) to synthesize photorealistic videos given a sequence of semantic label maps, sketches or human pose as an input.
During training, the network takes paired data as input, \eg sequences of a semantic label map and the corresponding RGB image sequence.
The network is constrained to preserve the content of the input sequence in the output video.

\begin{figure*}[t]
    \centering
    \includegraphics[width=.85\linewidth]{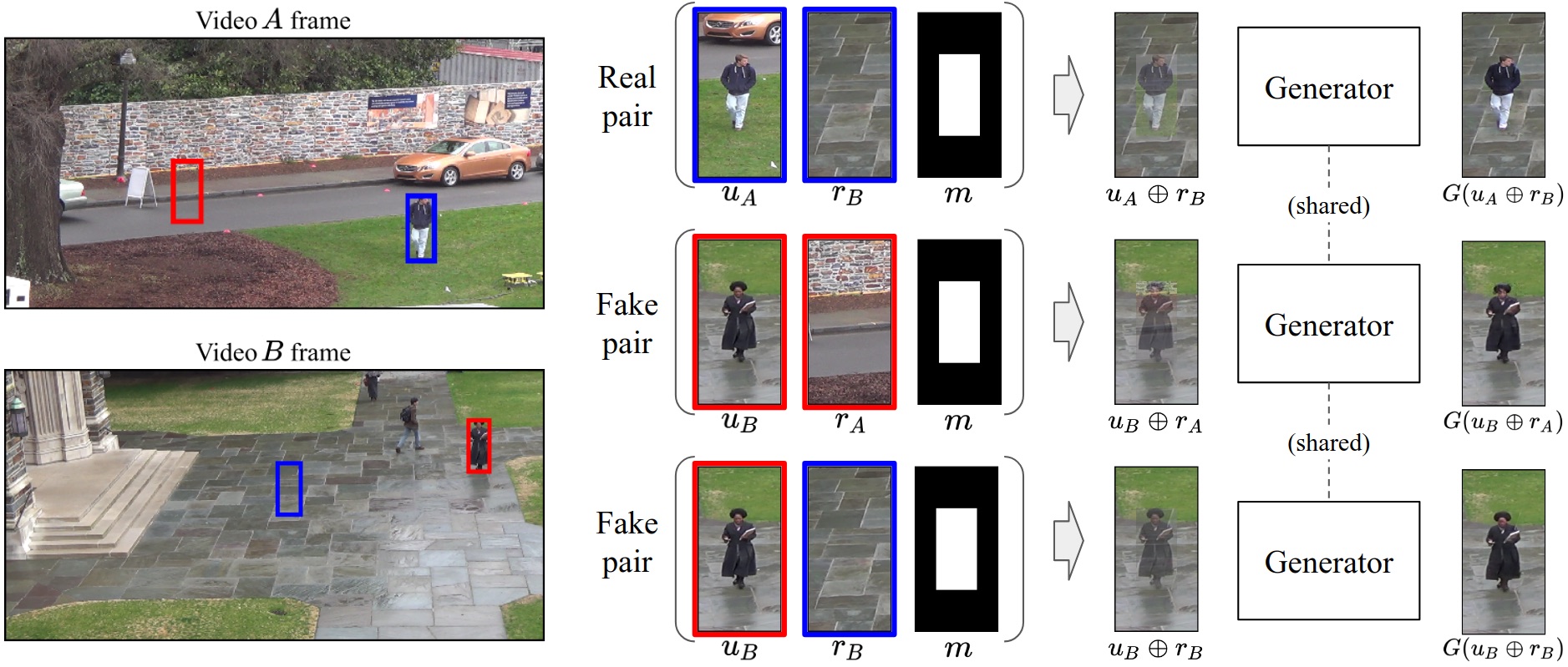}
    \caption{
    Main steps of the proposed algorithm for inserting an object into an image.
    Given the video $A$ and $B$, we crop objects and backgrounds from each video and synthesize blended images $u_A\oplus r_B$, $u_B\oplus r_A$, and $u_B\oplus r_B$.
    By learning how to reconstruct $u_B$ from $u_B\oplus r_A$ and $u_B\oplus r_B$, we can guide the network to insert $u_A$ into the center of $r_B$.
    In addition to reconstruction losses from fake pairs, we have additional objective functions as described in (\ref{eq:img_all_loss}).
    }
\label{fig:insert_img}
\vspace{-4mm}
\end{figure*}

\begin{figure}[t]
    \centering
    \subfloat[$u_A$]{{\includegraphics[width=.19\linewidth]{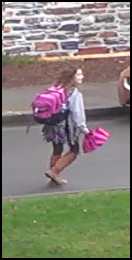} }}
    \subfloat[$r_B$]{{\includegraphics[width=.19\linewidth]{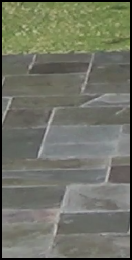} }}
    \subfloat[(\ref{eq:fake_pair_adv}) + (\ref{eq:fake_pair_recon})]{{\includegraphics[width=.19\linewidth]{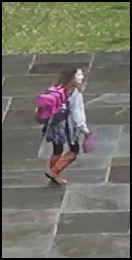} }}
    \subfloat[(\ref{eq:fake_pair_recon}) + (\ref{eq:both_pair_adv1})]{{\includegraphics[width=.19\linewidth]{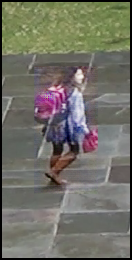} }}
    \subfloat[(\ref{eq:img_all_loss})]{{\includegraphics[width=.19\linewidth]{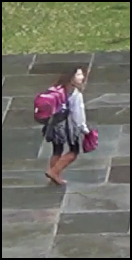} }}
    \hfill
    \caption{Object insertion results using different objective functions.
    Inputs are $u_A$ and $r_B$.}
    \label{fig:result_different_objective}
    \vspace{-4mm}
\end{figure}
\section{Proposed Algorithm}

In this work, we consider the problem where a user selects an object in video $A$ and wants to insert it at a desired location in video $B$.
We assume that each video has annotations for bounding boxes and IDs of objects at every frame.
From the bounding boxes of the selected object in $A$, we obtain a video $\mathbf{u}_A$ consisting of cropped images. 
The goal is to translate $\mathbf{u}_A$ to $\mathbf{v}_A$ so that the translated video is realistic when inserted into $B$. 
We first tackle this problem's image counterpart and then extend it to videos.

\subsection{Inserting images into images}
\label{sec:i2i}
Let $u_A$ denote a frame in $\mathbf{u}_A$ which will be inserted into a user-defined region $r_B$ in $B$.
We train a generator network $G_I$ which takes $u_A$ and $r_B$ as inputs to render an output $v_A$.
Note that this is different from existing image-to-image translation tasks~\cite{huang2018multimodal,isola2017image,liu2016unsupervised,CycleGAN2017,zhu2017toward} since they aim to preserve the content of an input image while changing it to different attributes or styles, \eg a semantic map is translated to RGB images that have the same semantic layout.
In contrast, we need to translate two different images into a single image while learning which part of the content in each image should be preserved.

One challenging issue is that we do not have a training tuple $(u_A, r_B, v_A)$.
To address this issue, we first cast the problem as a conditional image in-painting task.
More specifically, we corrupt $r_B$ by blending $u_A$ using pixel-wise multiplications with a fixed binary mask $m$, \ie $u_A \oplus r_B = u_Am/2 + r_B(1-m/2)$, as shown in the Figure~\ref{fig:insert_img}.
Then, the generator learns a mapping $G_I\colon(u_A\oplus r_B)\to v_A$ to synthesize realistic $v_A$.
To this end, the generator learns how to render the object while suppressing mismatched backgrounds based on the context of surrounding non-blended regions.
The key advantage of this formulation is that it is easy to synthesize fake training pairs that are similar to $(u_A\oplus r_B, v_A)$.

In this paper, we propose two types of fake pairs $(u_B\oplus r_A, u_B)$ and $(u_B\oplus r_B, u_B)$ to learn object insertion.
The intuition behind it is that these pairs contain two separate tasks that the generator has to perform during inference: rendering consistent backgrounds based on the context, and recovering the object region overlapped with $r_B$.
We design two objective functions for fake pairs using $G_I$ and an image discriminator $D_I$.
First,
\begin{equation}\label{eq:fake_pair_adv}
\begin{split}
  & \mathcal{L}_{\mathcal{A}}^{fake}(G_I, D_I) = \mathbb{E}_{(u_B,r_A)}[\log D_I(u_B, u_B \oplus r_A)] \\
  & \ + \mathbb{E}_{(u_B,r_B)}[\log D_I(u_B, u_B \oplus r_B)] \\
  & \ + \mathbb{E}_{(u_B,r_A)}[\log(1-D_I(G_I(u_B \oplus r_A), u_B \oplus r_A))] \\
  & \ + \mathbb{E}_{(u_B,r_B)}[\log(1-D_I(G_I(u_B \oplus r_B), u_B \oplus r_B))],
\end{split}
\end{equation}
is a conditional adversarial loss to make the reconstructed image sharper and realistic\footnote{\footnotesize
We denote $\mathbb{E}_{(\cdot)} \triangleq \mathbb{E}_{(\cdot)\sim p_{data}(\cdot)}$ for notational simplicity.}.
Second,
\begin{equation}\label{eq:fake_pair_recon}
    \mathcal{L}_{\mathcal{R}}(G_I) = \|u_B - G_I(u_B\oplus r_A)\| + \|u_B - G_I(u_B\oplus r_B)\|,
\end{equation}
is a content loss to reconstruct $u_B$.

We present results on the real pair using a network trained with fake pairs in Figure~\ref{fig:result_different_objective}(c).
Although some parts are blurry, the overall shape and appearance of inserted objects are preserved.
In addition, most of the background pixels from $A$ are removed and replaced by $r_B$, showing that fake pairs provide meaningful signals to the network to insert unseen objects.
Thus, we expect that the network can be trained well with both of real and fake pairs.
We update the adversarial loss to consider real pairs as follows:
\begin{equation}\label{eq:both_pair_adv1}
\begin{split}
  & \mathcal{L}_{\mathcal{A}}(G_I,D_I) = \mathcal{L}_{\mathcal{A}}^{fake}(G_I, D_I) \\
  & \ + \mathbb{E}_{(u_A,r_B)}[\log(1-D_I(G_I(u_A \oplus r_B), u_A \oplus r_B))].
\end{split}
\end{equation}

However, as shown in Figure~\ref{fig:result_different_objective}(d), the synthesized results become unstable when we naively train the network using (\ref{eq:fake_pair_recon}) and (\ref{eq:both_pair_adv1}).
We attribute this to different distributions of the fake pair and real pair.
Although their similar distributions make it possible to generalize the network to unseen images, when the network actually learns with both pair types, it is able to distinguish between them, thus limiting generalization.
We address this issue by making it more difficult for the network to distinguish these pairs. 
In particular, we make it uncertain about whether the input is sampled from the fake pair or real pair.
To this end, we add a discriminator $D_E$ that aims to distinguish the input type based on its embedded vector as follows:
\begin{equation}\label{eq:embed_adv}
\begin{split}
  \mathcal{L}_{\mathcal{A}}(G_I, D_E) &= \mathbb{E}_{(u_B,r_A)}[\log D_E(e_{u_B \oplus r_A})] \\
  & + \mathbb{E}_{(u_B,r_B)}[\log D_E(e_{u_B \oplus r_B})] \\
  & + \mathbb{E}_{(u_A,r_B)}[\log (1-D_E(e_{u_A \oplus r_B}))],
\end{split}
\end{equation}
where $e_x$ denotes an embedded vector from the encoder in $G_I$ with an input $x$.
The encoder is trained to fool the discriminator by embedding the fake pair and real pair into the same space.
This embedding vector is fed to discriminators as a conditional input.
%
We tile the vector to the same size of the input image and concatenate them to the input channel.
The objective function $\mathcal{L}_{\mathcal{A}}(G_I,D_I)$ is modified as follows:
\begin{equation}\label{eq:both_pair_adv2}
\begin{split}
  & \mathcal{L}_{\mathcal{A}}(G_I, D_I) = \mathbb{E}_{(u_B,r_A)}[\log D_I(u_B, e_{u_B \oplus r_A})] \\
  & \ + \mathbb{E}_{(u_B,r_B)}[\log D_I(u_B, e_{u_B \oplus r_B})] \\
  & \ + \mathbb{E}_{(u_B,r_A)}[\log(1-D_I(G_I(u_B \oplus r_A), e_{u_B \oplus r_A}))] \\
  & \ + \mathbb{E}_{(u_B,r_B)}[\log(1-D_I(G_I(u_B \oplus r_B), e_{u_B \oplus r_B}))] \\
  & \ + \mathbb{E}_{(u_A,r_B)}[\log(1-D_I(G_I(u_A \oplus r_B), e_{u_A \oplus r_B}))].
\end{split}
\end{equation}

Finally, the overall objective function for object insertion on the image domain is formulated as follows:
\small
\begin{equation}\label{eq:img_all_loss}
    \mathcal{L}(G_I, D_I, D_E) = \mathcal{L}_{\mathcal{A}}(G_I, D_I) + \mathcal{L}_{\mathcal{A}}(G_I, D_E) + \mathcal{L}_{\mathcal{R}}(G_I).
\end{equation}
\normalsize
Figure~\ref{fig:result_different_objective}(e) shows that the inserted objects using the loss function in~\eqref{eq:img_all_loss} are sharp and realistic.

\begin{figure}[t!]
    \centering
    \includegraphics[width=.9\linewidth]{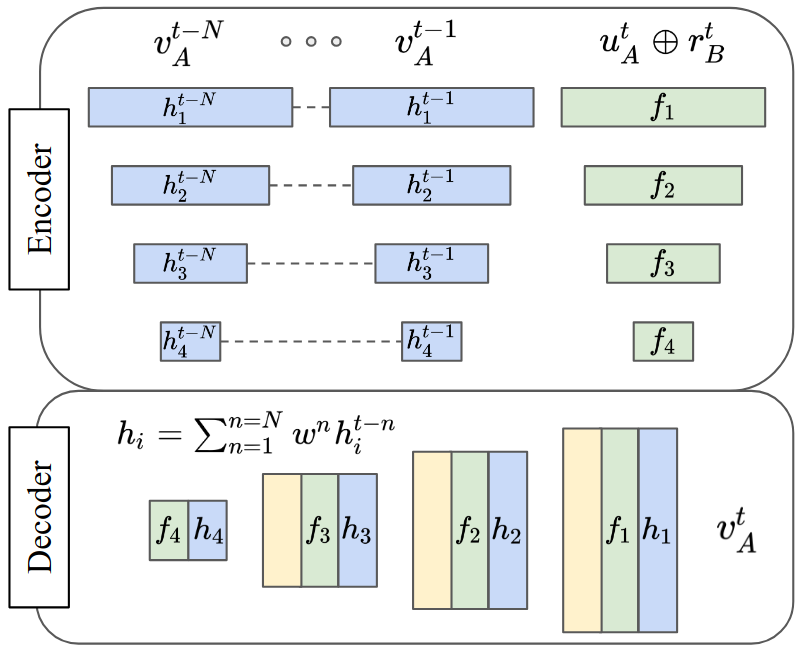}
    \caption{Network structure of the video insertion network $G_V$.
    As an illustrative example, we show the case where the number of layers is four.
    The network takes previous frames $(v_A^{t-N},\dots,v_A^{t-1})$ and a blended image $u_A^t\oplus r_B^t$ as an input to render $v_A^t$.
    Each square denotes a layer in the network.
    The dashed lines indicates shared weights, and layers next to each other represent channel concatenations.
    }
\label{fig:Gv}
\vspace{-4mm}
\end{figure}

\subsection{Inserting videos into videos}
In this section, we discuss how to extend the object insertion model from images to videos.
To this end, we make two major modifications.
First, when rendering the current frame, we also look up previous frames.
Second, we add a new term in the objective function to synthesize temporally consistent videos.

Let $G_V$ denote a video generator that learns a mapping $G_V\colon (\mathbf{u}_A\oplus \mathbf{r}_B) \to \mathbf{v}_A$\footnote{\footnotesize
We denote $(\mathbf{u}_A\oplus \mathbf{r}_B)$ as a sequence of blended inputs $((u_A^{1}\oplus r_B^{1},\dots,(u_A^{T}\oplus r_B^{T}))$ where $T$ is the number of frames.}.
One simple mapping is to apply $G_I$ for each frame.
However, as the mapping of a frame is independent from neighboring frames, the resulting sequence becomes temporally inconsistent.
Therefore, we let $G_V$ to additionally look up $N$ previous frames while synthesizing each frame from the blended input.
This Markov assumption is useful for generating long sequence videos \cite{wang2018video}.
Figure~\ref{fig:Gv} shows the proposed U-net \cite{ronneberger2015u} style encoder-decoder network architecture.
If the network operates without blue layers, which correspond to the feature maps of previous frames, then it is identical to $G_I$ in Section~\ref{sec:i2i}.
The network encodes all previous frames using a shared encoder.
Then, the feature map is linearly combined with a scalar weight $w^n$ which represents the importance of each frame.
We use $N=2$ and $w^1=w^2=0.5$ for experiments in this work. 

To learn $G_V$, we calculate an error signal for the generated sequence using the following objective function:
\small
\begin{equation}\label{eq:vid_all_loss}
\begin{split}
    \mathcal{L}(G_V, D_I, D_V, D_E) & = \mathcal{L}_{\mathcal{A}}(G_V, D_I) + \mathcal{L}_{\mathcal{A}}(G_V,D_V)\\
    & + \mathcal{L}_{\mathcal{A}}(G_V, D_E) + \mathcal{L}_{\mathcal{R}}(G_V),
\end{split}
\end{equation}
\normalsize
where $D_V$ is a video discriminator.
The first term is defined similarly to (\ref{eq:both_pair_adv2}) while we select a random frame from the generated sequence to calculate the loss; this term focuses on the realism of the selected frame.
The second term assesses the rendered sequence as follows:
\small
\begin{equation}\label{eq:GvDv}
\begin{split}
  & \mathcal{L}_{\mathcal{A}}(G_V, D_V) = \mathbb{E}_{(\mathbf{u}_B,\mathbf{r}_A)}[\log D_V(\mathbf{u}_B, e_{\mathbf{u}_B \oplus \mathbf{r}_A})] \\
  & \ + \mathbb{E}_{(\mathbf{u}_B,\mathbf{r}_B)}[\log D_V(\mathbf{u}_B, e_{\mathbf{u}_B \oplus \mathbf{r}_B})] \\
  & \ + \mathbb{E}_{(\mathbf{u}_B,\mathbf{r}_A)}[\log(1-D_V(G_V(\mathbf{u}_B \oplus \mathbf{r}_A), e_{\mathbf{u}_B \oplus \mathbf{r}_A}))] \\
  & \ + \mathbb{E}_{(\mathbf{u}_B,\mathbf{r}_B)}[\log(1-D_V(G_V(\mathbf{u}_B \oplus \mathbf{r}_B), e_{\mathbf{u}_B \oplus \mathbf{r}_B}))] \\
  & \ + \mathbb{E}_{(\mathbf{u}_A,\mathbf{r}_B)}[\log(1-D_V(G_V(\mathbf{u}_A \oplus \mathbf{r}_B), e_{\mathbf{u}_A \oplus \mathbf{r}_B}))].
\end{split}
\end{equation}
\normalsize
The third and fourth terms are defined similarly to (\ref{eq:embed_adv}) and (\ref{eq:fake_pair_recon}), respectively.

In addition, while training the network, we observe that the predicted frame $v_A^t$ heavily relies on the previous frames rather than the current input.
The main reason is that the current input is corrupted by a blending $u_A^t\oplus r_B^t$ which makes it more difficult to process.
Therefore, instead of learning to recover the current frame, the network gradually ignores the current input and depends more on the previous frame.
It is a critical problem when generating long videos as the error from the previous frame is accumulated.
As a result, the generated sequence contains severe artifacts after a number of frames.
To address this issue, we degrade previous frames as well using random noise before render the current frame.
By blocking this easy cheating route, the network has to learn semantic relationships between the two inputs instead of relying on one side.
It makes the network significantly stable during training.

\section{Experimental Results}
\label{sec:experiments}
We evaluate our method on the multi-target tracking or person re-identification databases such as the DukeMTMC \cite{ristani2016MTMC}, TownCenter \cite{benfold2011stable}, and UA-DETRAC \cite{wen2015ua} to show applicability of our algorithm on real-world examples.
These datasets record challenging scenarios where pedestrians or cars move naturally.
We split 20\% of the data as a test set and present experimental results on the test set.
%
Additional results, including sample generated videos and a user study, are included in the supplementary material.

\vspace{-4mm}
\paragraph{Implementation details.}
For all experiments, the network architecture, parameters, and initialization are similar to DCGAN~\cite{radford2015unsupervised}.
%
We use transposed convolutional layers with 64 as a base number of filters for both of the generator and discriminator.
The batch size is set to 1 and instance normalization is used instead of batch normalization.
Input videos are resized to $1024\times2048$ pixels.
We crop $u_{(\cdot)}$ and $r_{(\cdot)}$ from the video and resize to $256\times128$ pixels.
Then, we render an object on the $256\times128$ pixels patch.
It is transformed to $512\times256$ pixels image or video for visualization.
For each iteration, we pick a random location in $A$ to put a new object since we want to cover various location and size input of a user.

\begin{figure*}[t]
    \centering
    \subfloat[$u_A$]{{\includegraphics[width=.1\linewidth]{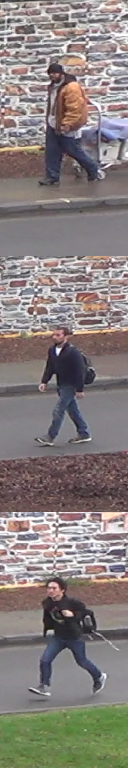} }}
    \subfloat[$r_B$]{{\includegraphics[width=.1\linewidth]{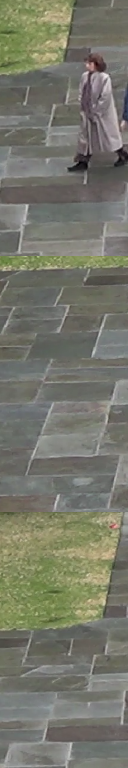} }}
    \subfloat[\cite{deeplabv3plus2018}]{{\includegraphics[width=.1\linewidth]{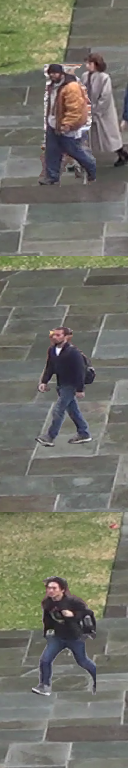} }}
    \subfloat[\cite{deeplabv3plus2018} + \cite{perez2003poisson}]{{\includegraphics[width=.1\linewidth]{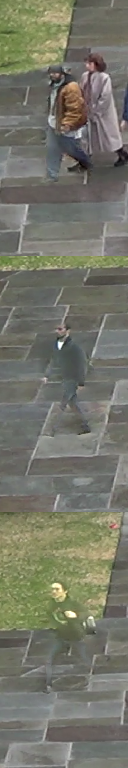} }}
    \subfloat[(\ref{eq:base_gan1})]{{\includegraphics[width=.1\linewidth]{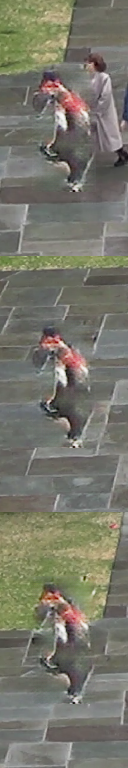} }}
    \subfloat[(\ref{eq:base_gan2})]{{\includegraphics[width=.1\linewidth]{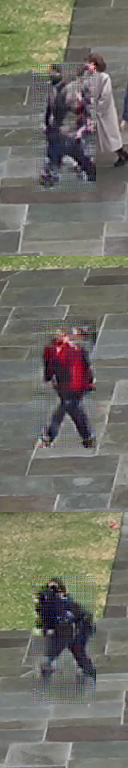} }}
    \subfloat[(\ref{eq:base_gan3})]{{\includegraphics[width=.1\linewidth]{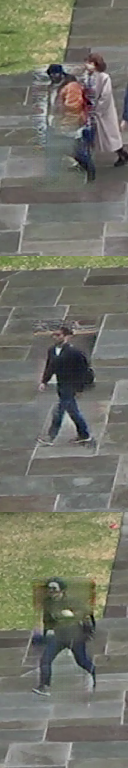} }}
    \subfloat[(\ref{eq:base_gan4})]{{\includegraphics[width=.1\linewidth]{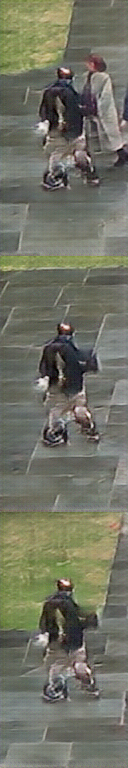} }}
    \subfloat[Ours (\ref{eq:img_all_loss})]{{\includegraphics[width=.1\linewidth]{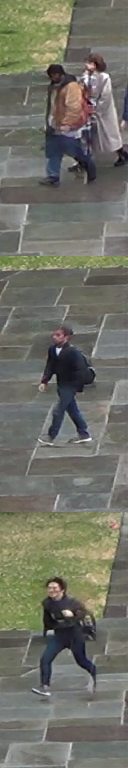} }}
    \hfill
    \caption{Object insertion results for different baseline models given input $u_A$ and $r_B$.}
    \label{fig:baseline}
    \vspace{-4mm}
\end{figure*}

\vspace{-4mm}
\paragraph{Baseline models and qualitative evaluations.}
As the problem introduced in this paper is a new problem, we design strong baselines for performance evaluation. 

For object insertion in images, we present six baseline models.
First, we apply the state-of-the-art semantic segmentation algorithm \cite{deeplabv3plus2018} to segment the interested object region in video $A$, \eg a pedestrian in the DukeMTMC dataset.
Then, object pixels are copied to a region in video $B$ using the predicted segmentation mask as shown in Figure~\ref{fig:baseline}(c).
However, the predicted segmentation mask is inaccurate due to the complex background and articulated human pose.
Therefore, some parts of the object are often missing and undesired background pixels from video $A$ are included in the synthesized frame.
In addition, the brightness of the inserted pixels does not match with surrounding pixels in video $B$.
Second, we apply the Poisson blending \cite{perez2003poisson} method to the predicted object mask as shown in Figure~\ref{fig:baseline}(d).
Although the boundary of object becomes smoother, the blended image still contains artifacts.
In addition, the results depend on the performance of the segmentation algorithm.

Third, we design four GAN-based methods.
One naive approach focuses on synthesizing realistic example using the following objective function:
\small
\begin{equation}\label{eq:base_gan1}
\begin{split}
  \mathcal{L}_{\mathcal{A}}^{base}(G, D) &= \mathbb{E}_{u_B}[\log D(u_B)] \\
  & + \mathbb{E}_{(u_A,r_B)}[\log D(G(u_A \oplus r_B))].
\end{split}
\end{equation}
\normalsize
In this case, the generator easily collapses as it is not guided to preserve the content of the input object as shown in Figure~\ref{fig:baseline}(e).
To alleviate this issue, we add an objective function that checks the content in the generated image, \eg a pixel-wise reconstruction loss or the perceptual loss~\cite{gatys2016image} as shown in Figure~\ref{fig:baseline}(f) and Figure~\ref{fig:baseline}(g).
The objective functions are defined as follows:
\small
\begin{equation}\label{eq:base_gan2}
\begin{split}
  \mathcal{L}_{pixel}^{base}(G, D) &= \mathcal{L}_{\mathcal{A}}^{base}(G,D) + \|u_Am - v_Am\|,
\end{split}
\end{equation}
\begin{equation}\label{eq:base_gan3}
\begin{split}
  & \mathcal{L}_{perceptual}^{base}(G, D) = \mathcal{L}_{\mathcal{A}}^{base}(G,D) \\
  & \qquad\qquad + \sum_l \frac{1}{C_l H_l W_l}\|\phi_l(u_Am)-\phi_l(v_Am)\|^2_2,
\end{split}
\end{equation}
\normalsize
where $\phi_l$ is $l$-th activation map of the VGG19 network \cite{simonyan15} with a shape of $C_l\times H_l\times W_l$.
We use activation maps of relu2\_2 and relu3\_3 layers of the VGG19 network which is pre-trained on the ImageNet dataset~\cite{russakovsky2015imagenet} to calculate the perceptual loss.
The main limitation of these approaches is that the network is trained to preserve all pixels around the object in $u_A$.
As a result, a large number of undesired background pixels appear in $v_A$.

The final baseline model uses the cycle consistency loss~\cite{CycleGAN2017} which has been used to train networks with unpaired training data.
For the cyclic loss, we learn two mapping functions $G\colon (u_A, r_B) \to v_A$ and $F\colon (u_B, r_A) \to v_B$.
By taking the conditional inputs into account, the objective function is defined by:
\small
\begin{equation}\label{eq:base_gan4}
\begin{split}
  & \mathcal{L}_{cyc}^{base}(G, F, D_A, D_B) = \mathcal{L}_{\mathcal{A}}(G,D_B) + \mathcal{L}_{\mathcal{A}}(F,D_A) \\
  & \ + \mathbb{E}_{(u_A,r_B)}[\|F(G(u_A,r_B),u_A(1-m))-u_A\|_1] \\
  & \ + \mathbb{E}_{(u_B,r_A)}[\|G(F(u_B,r_A),u_B(1-m))-u_B\|_1] \\
  & \ + \mathbb{E}_{(u_A,r_B)}[\|G(u_A,r_B)(1-m)-r_B(1-m)\|_1] \\
  & \ + \mathbb{E}_{(u_B,r_A)}[\|F(u_B,r_A)(1-m)-r_A(1-m)\|_1],
\end{split}
\end{equation}
\normalsize
where $D_A$ and $D_B$ are discriminators for each video and $\mathcal{L}_{\mathcal{A}}(G,D_B)$ and $\mathcal{L}_{\mathcal{A}}(F,D_A)$ are typical adversarial losses.
The last two terms are added to force the network to insert an object at a given $r_A$ or $r_B$.
Although the formulation has the potential to learn unpaired mappings, it still cannot guide the network to preserve the same object while translating images as shown in Figure~\ref{fig:baseline}(h).
In addition, we observe that this makes the network unstable during training.
In contrast, the proposed algorithm inserts an object with its sharp shape and renders less noisy background pixels as shown in Figure~\ref{fig:baseline}(i).

For video object insertion, we consider two baseline models.
First, frames are synthesized without using previous frames.
As the model only processes the current frame as an input, the overall video may contain flickering or inconsistent content.
Second, a video is generated without injecting noise into previous frames.
%
In such cases, as small errors in each frame accumulate over frame, the synthesized images are likely noisy.

\begin{figure*}[t]
    \centering
    \href{https://youtu.be/-lL8zPYYNV4}{\includegraphics[width=0.86\linewidth]{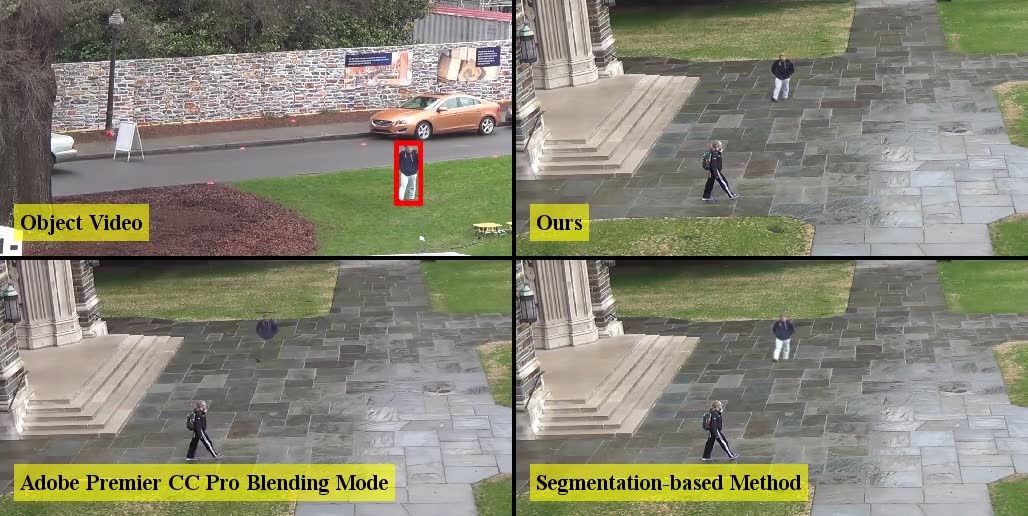}}
    \caption{
    Inserting a pedestrian video on the DukeMTMC dataset.
    \textcolor{red}{Click the image to play the video.}
    }
\label{fig:vid1}
\vspace{-4mm}
\end{figure*}

\begin{figure}[t]
    \centering
    \href{https://youtu.be/iOJcp-JubWA}{\includegraphics[width=0.95\linewidth]{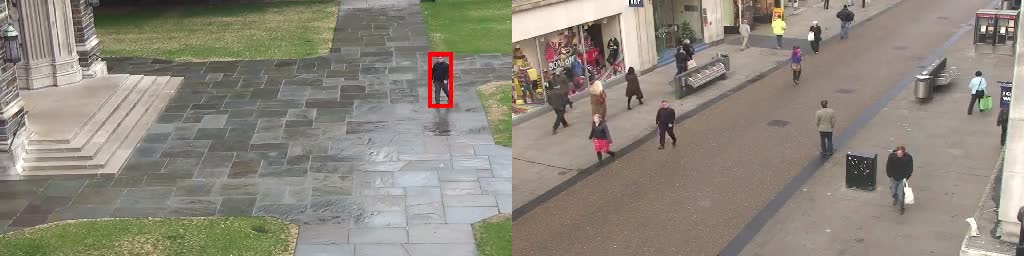}}
    \href{https://youtu.be/v38b-uD0t3o}{\includegraphics[width=0.95\linewidth]{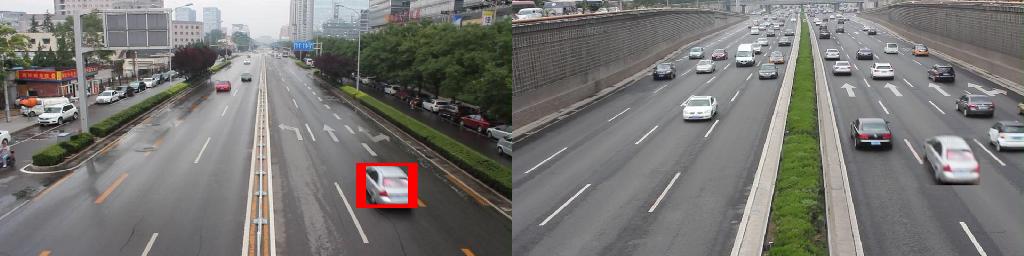}}
    \caption{
    Results of cross dataset pedestrian insertion (from the DukeMTMC dataset to TownCenter dataset) and a car video insertion on the UA-DETRAC dataset.
    }
\label{fig:vid2}
\vspace{-5mm}
\end{figure}



Figure~\ref{fig:vid1} shows video object insertion results with baseline comparisons.
We use an automatic blending mode of a commercial video editing software (Adobe Premier CC Pro) as one baseline.
The other baseline uses DeepLabv3+ \cite{deeplabv3plus2018} to copy and paste the predicted segment along frames.
It shows that the proposed algorithm can synthesize more realistic videos than other baseline methods.
In addition, as shown in Figure~\ref{fig:vid2}, our algorithm is capable of inserting videos across databases and different objects such as a car.

\begin{table}[t!]
    \centering
    \caption{Recall of the state-of-the-art object detector \cite{redmon2018yolov3} on the DukeMTMC database. B1: Adobe Premiere blending mode. B2: Segmentation-based composition \cite{deeplabv3plus2018}.}
    \begin{tabular}{cccccccc}
    \toprule
     Method & B1 & B2 &
     (\ref{eq:base_gan1}) &
     (\ref{eq:base_gan2}) &
     (\ref{eq:base_gan3}) &
     Our \\
     \midrule
     Recall & 0.39 & 0.76 & 0.73 & 0.80 & 0.78 & \textbf{0.86} \\
    \bottomrule
    \end{tabular}
    \label{tab:det_recall}
    \vspace{-3mm}
\end{table}

\begin{table}[t!]
    \centering
    \caption{Object insertion score on the DukeMTMC dataset. B1: Adobe Premiere blending mode.
    }
    \begin{tabular}{cccccc}
    \toprule
     Method & B1 &
     (\ref{eq:base_gan1}) &
     (\ref{eq:base_gan2}) &
     (\ref{eq:base_gan3}) &
     Our \\
     \midrule
     Precision & 0.32 & 0.61 & 0.70 & 0.76 & \textbf{0.85} \\
     Recall & 0.28 & 0.26 & 0.47 & 0.61 & \textbf{0.72} \\
     \midrule
     OIS & 0.30 & 0.36 & 0.56 & 0.68 & \textbf{0.78} \\
    \bottomrule
    \end{tabular}
    \label{tab:seg_ois}
    \vspace{-4mm}
\end{table}

\vspace{-4mm}
\paragraph{Quantitative evaluations.}
To quantify the realism of the inserted object, an object detector is often 
used to locate the inserted object \cite{lee2018context,ouyang2018pedestrian,chien2017detecting}.
The premise is that a detector is likely to locate only well-inserted objects since state-of-the-art methods take both of the object and its surrounding context into account.
We use the YOLOv3 detector \cite{redmon2018yolov3} to determine whether it can correctly detect the inserted object or not.
We fix the detection threshold and measure the recall of the detector by calculating the intersection over union (IoU) between the inserted object and detected bounding boxes, using an IoU threshold of 0.5.
Table~\ref{tab:det_recall} shows the average recall using a network trained with five different iterations.
For each experiment, we sample one thousand images at random.
It shows that the proposed algorithm achieves the highest recall value on average.
In addition, we accidentally found an interesting corner case of this experiment.
While (\ref{eq:base_gan1}) generates non-realistic images in a similar mode as shown in Figure~\ref{fig:baseline}(e), this method once achieves the highest recall value.
It reveals one limitation of assessing the synthesized image using a detector, \ie, if a trained detector mistakenly returns positive detection result for a non-realistic fake image, then it is highly likely that other non-realistic images in the same mode will be detected as positive samples as well.

While detection results give an idea of how realistic (or at least, detectable) the inserted object is, it does not indicate pixel-level accuracy of the object insertion, \ie whether the object pixels in the input are preserved in the output.
To this end, we introduce a new metric based on pixel-level precision and recall for object insertion.
Given a semantic segmentation algorithm, let $s_A$ denote a binary segmentation mask of the input object image.
Also let $s_\Delta$ be a binary mask where $s_\Delta(i,j)=1$ when $v_A(i,j)$ is closer to $u_A(i,j)$ than $r_B(i,j)$.
Thus, $s_\Delta$ represents pixel locations of the inserted object.
We then define the precision $P$, recall $R$, and object insertion score (OIS) as follows:
\begin{equation}\label{eq:precision_recall}
  P = \frac{|s_A\odot s_\Delta|}{|s_\Delta|}, \
  R = \frac{|s_A\odot s_\Delta|}{|s_A|}, \
  \text{OIS} = 2\frac{PR}{P+R},
\end{equation}
where $\odot$ is an element-wise multiplication, $|s|$ is an area of non-zero region in $s$, and OIS is defined using the $F_1$ score.
We calculate the score based on randomly generated one thousand samples and segmentation masks are obtained by the DeepLabv3+ \cite{deeplabv3plus2018} method.
Table~\ref{tab:seg_ois} shows that the proposed algorithm achieves the highest OIS against other baseline algorithms.
We also note that the OIS of the baseline model based on (\ref{eq:base_gan1}) is the lowest. 

In order to show potential application for data augmentation, we train a detector using synthesized objects by our algorithm.
We detect pedestrians on the DukeMTMC dataset using YOLOv3 initialized on the ImageNet.
For training and evaluation, we pick 100 and 1,000 frames at random from the video of camera 5 in the dataset.
In addition, 3,000 frames are augmented by inserting pedestrians from camera 1.
It boosts mAP from 53.1\% to 68.3\%.


\section{Conclusion}
In this paper, we have introduced an algorithm to a new problem: manipulating a given video by inserting other videos into it.
It is a challenging task as it is inherently an unsupervised (unpaired) problem.
Unlike existing approaches, we propose an algorithm that converts the problem to a paired problem by synthesizing fake training pairs and corresponding loss functions.
We conducted experiments on real-world videos and demonstrated that the proposed algorithm is able to render long realistic videos with a given object video inserted.
As a future work, it is interesting to make the inserted object interact with the new video, \eg, path navigation or occlusion handling. 

\clearpage 

\title{Inserting Videos into Videos \\ Supplementary Material}

\author{\vspace{-5ex}}

\maketitle

In this supplementary material, we describe additional experimental results.

\section{Quantitative Results}
\begin{figure}[t]
    \centering
    \subfloat[]{\includegraphics[width=1\linewidth]{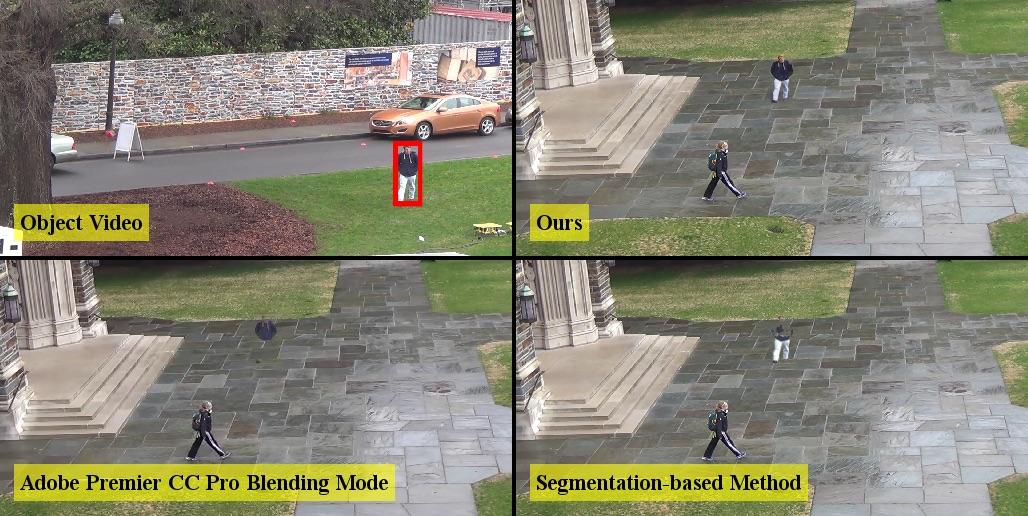}} \\
    \subfloat[]{\includegraphics[width=1\linewidth]{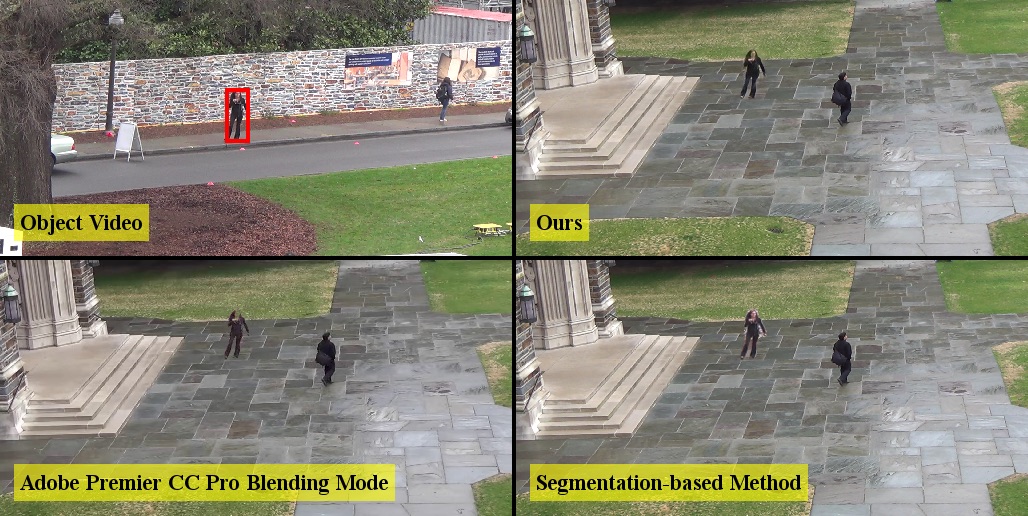}}
    \caption{
    Performance evaluation with baseline methods.
    Upper left: a frame from a video of a target object.
    Upper right: results of the proposed algorithm.
    Bottom left: automated blending results using a video editing software.
    Bottom right: results using a segmentation algorithm \cite{deeplabv3plus2018}.
    }
\label{fig:ex1}
\end{figure}

\begin{figure}[t]
    \centering
    \subfloat[]{\includegraphics[width=1\linewidth]{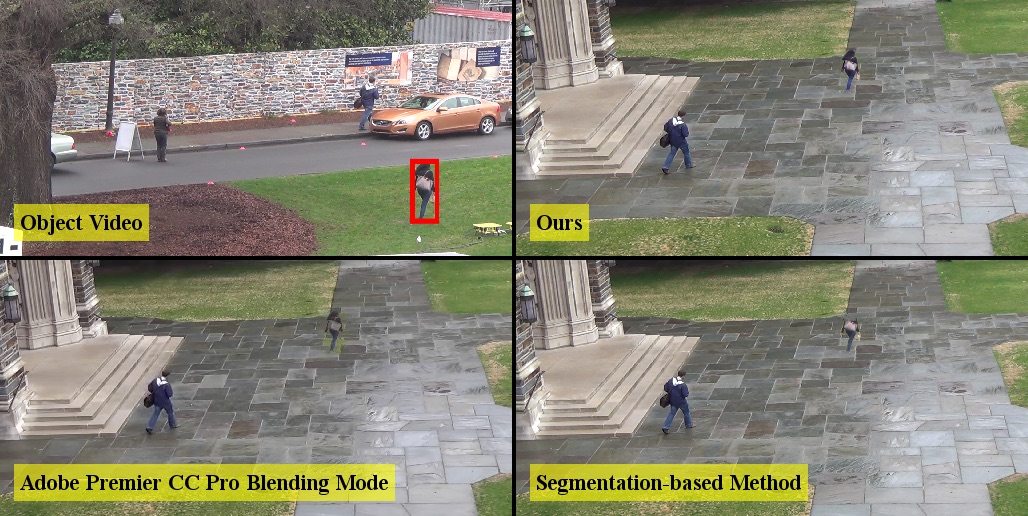}} \\
    \subfloat[]{\includegraphics[width=1\linewidth]{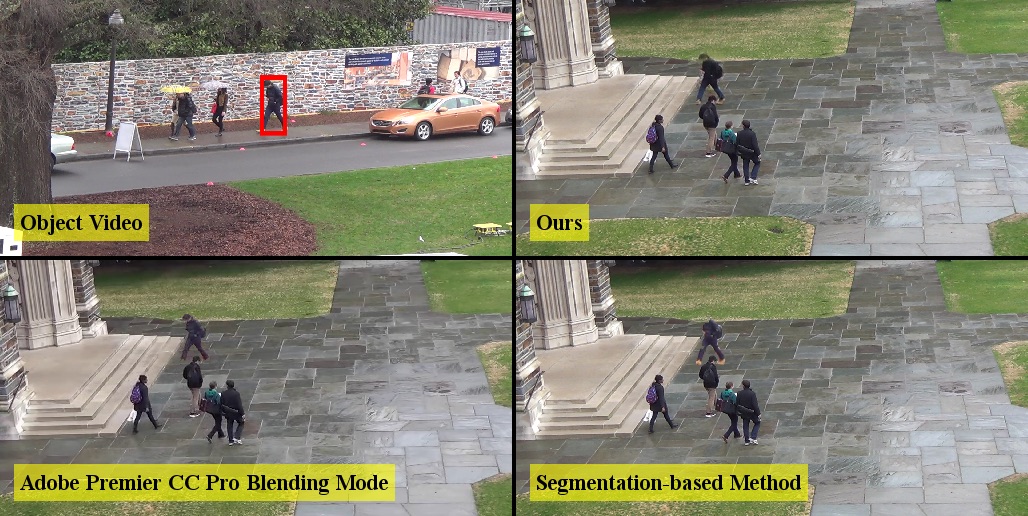}}
    \caption{
    Performance evaluation  with baseline methods.
    }
\label{fig:ex2}
\end{figure}

\begin{figure}[t]
    \centering
    \subfloat[]{\includegraphics[width=1\linewidth]{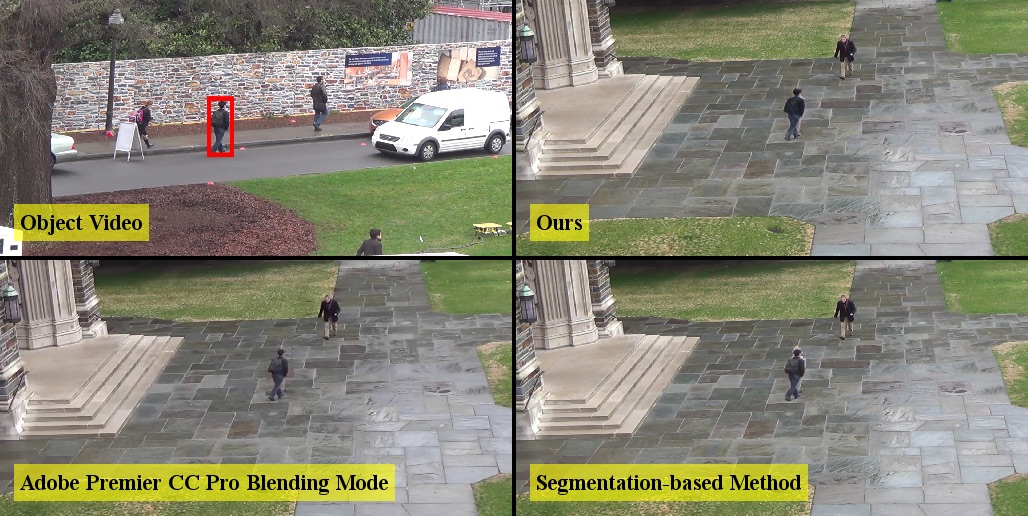}} \\
    \subfloat[]{\includegraphics[width=1\linewidth]{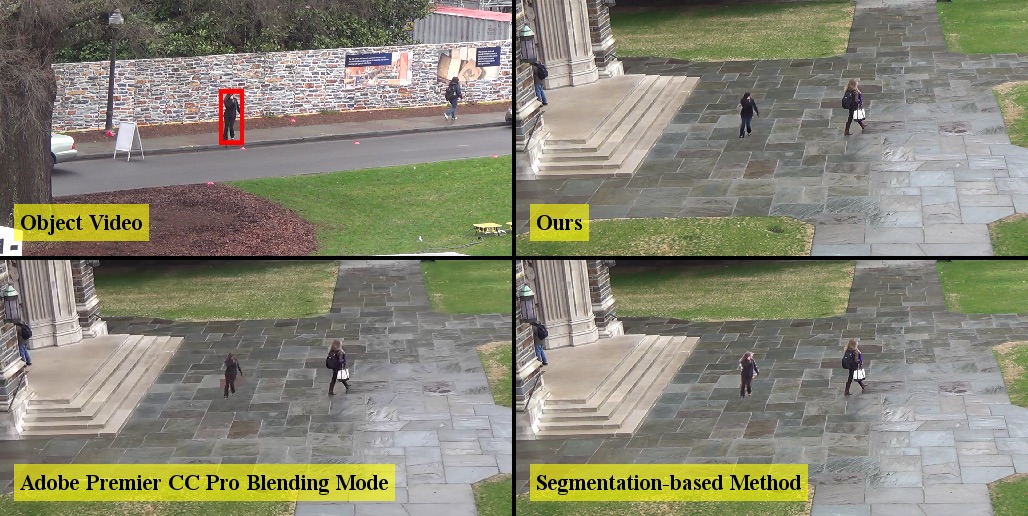}}
    \caption{
    Performance evaluation with baseline methods.
    }
\label{fig:ex3}
\end{figure}

\begin{figure}[t]
    \centering
    \subfloat[]{\includegraphics[width=1\linewidth]{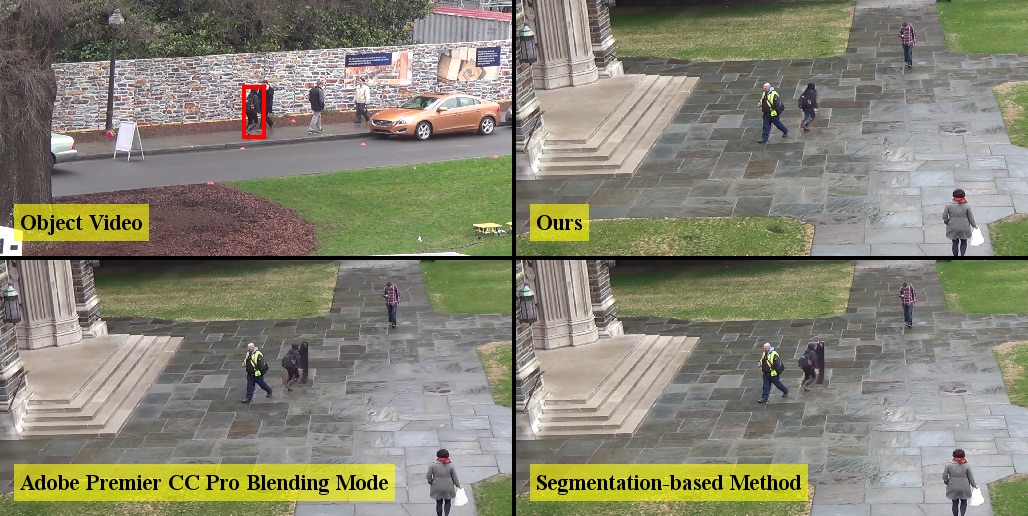}} \\
    \subfloat[]{\includegraphics[width=1\linewidth]{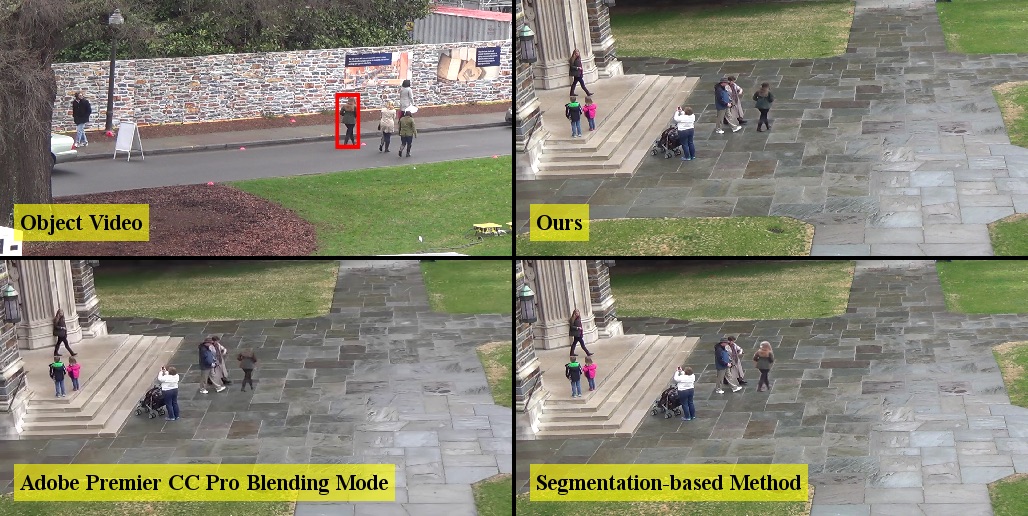}}
    \caption{
    Performance evaluation with baseline methods.
    }
\label{fig:ex4}
\end{figure}

\begin{figure}[t]
    \centering
    \subfloat[]{\includegraphics[width=1\linewidth]{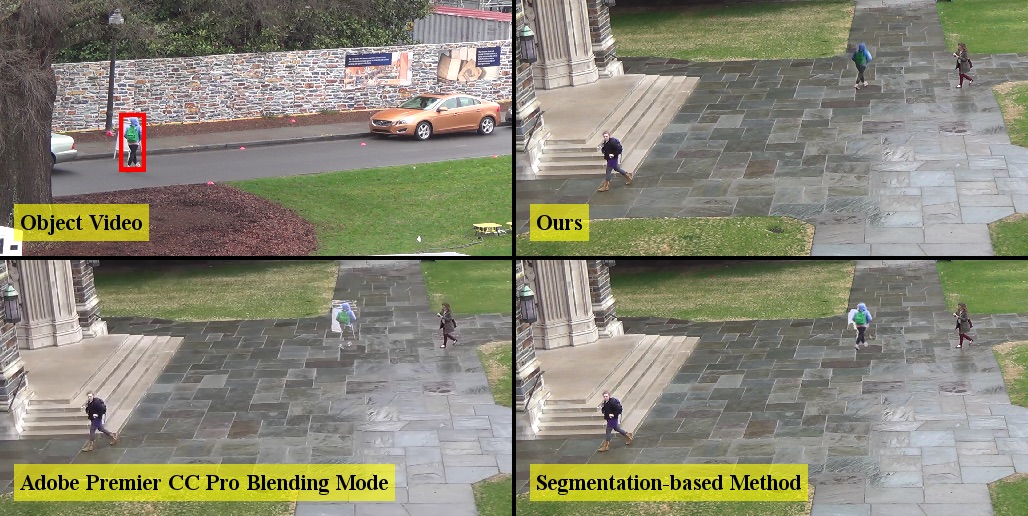}} \\
    \subfloat[]{\includegraphics[width=1\linewidth]{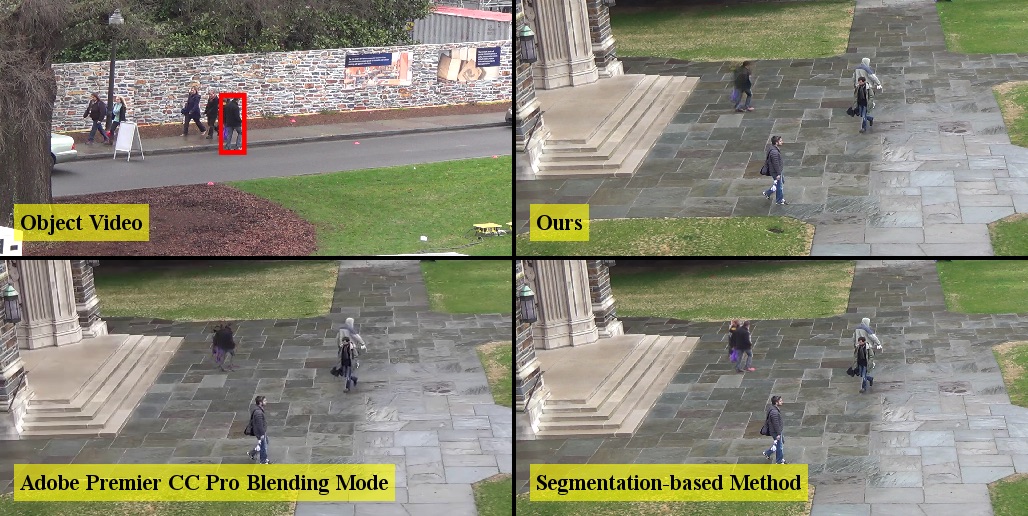}}
    \caption{
    Performance evaluation with baseline methods.
    The results present failure cases of our algorithm.
    }
\label{fig:fail}
\end{figure}

\begin{figure}[t]
    \centering
    \subfloat[]{\includegraphics[width=1\linewidth]{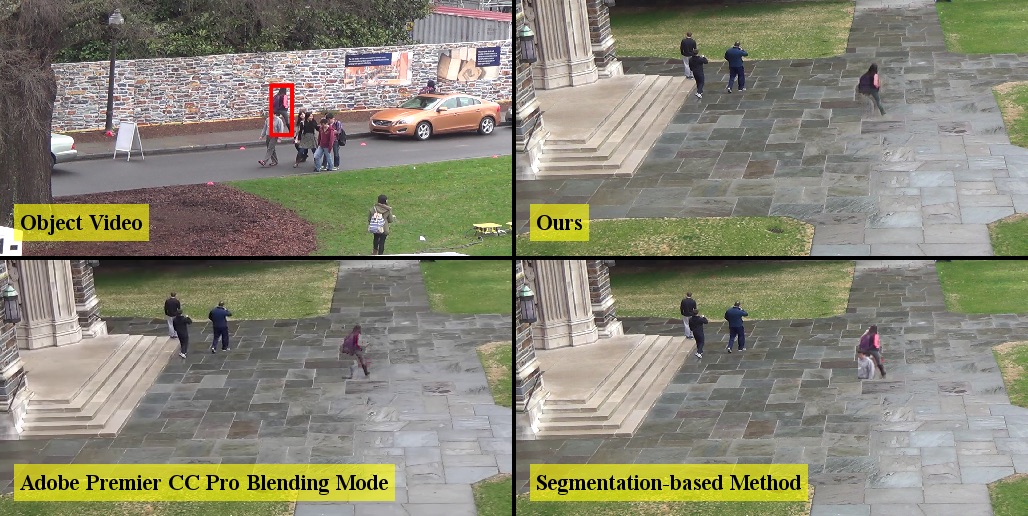}} \\
    \subfloat[]{\includegraphics[width=1\linewidth]{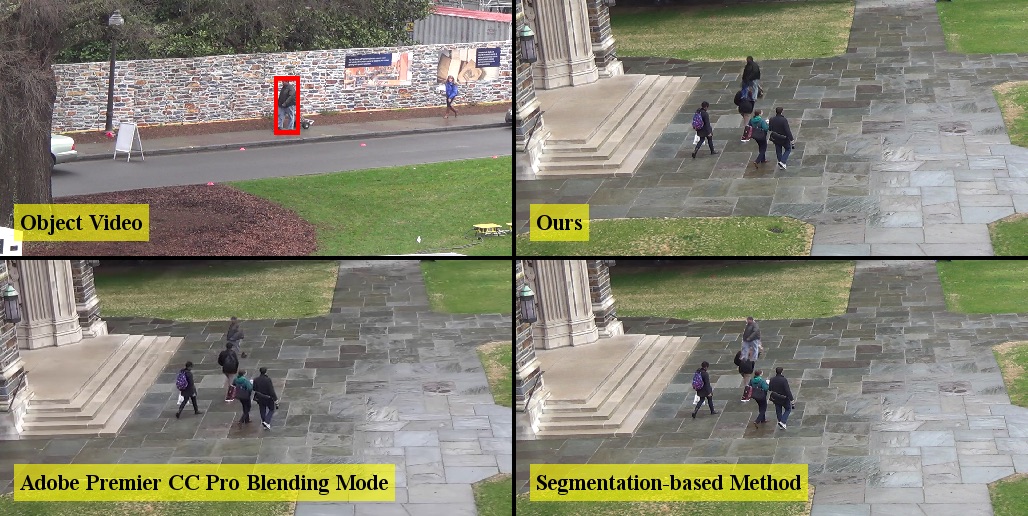}}
    \caption{
    Performance evaluation with baseline methods.
    The results show two different occlusion cases.
    }
\label{fig:occ}
\end{figure}

As the problem on inserting videos into video is new in the field, there are no existing methods that achieve this task.
Sample frames from videos are shown in Figure~\ref{fig:ex1} to Figure~\ref{fig:occ}.
For each figure, at the upper left corner of the footage, we display a frame from video $A$ that contains the target object marked in a red box.
Inserted objects into video $B$ using the proposed algorithm are presented at the upper right corner.
Rendering results of a video editing software, Adobe Premier Pro CC, is located at the bottom left corner as the first strong baseline method.
We use blending mode of the software to automatically overlay two videos.
The second strong baseline deployed at the bottom right corner is based on the state-of-the-art segmentation algorithm \cite{deeplabv3plus2018}.
It often segments the target object incorrectly, \ie, some parts are missing (Figure~\ref{fig:ex1}(a)) or backgrounds are included (Figure~\ref{fig:ex1}(b)).
Experimental results show that the proposed algorithm synthesizes more realistic videos in most cases.

We discuss our two different failure cases shown in Figure~\ref{fig:fail} and Figure~\ref{fig:occ}.
If the image patch of the target object contains different objects or rare backgrounds, then the synthesized object is less realistic as shown in Figure~\ref{fig:fail}.
This issue can be alleviated by collecting more data.
Occlusions caused by other pedestrians or objects in the scenes are another challenging case.
If the object is occluded in video $A$ as shown in Figure~\ref{fig:occ}(a), then ideally the algorithm has to infer the occluded part and infill the missing part.
In Figure~\ref{fig:occ}(b), the object has to be inserted behind an existing object in video $B$.
It is particularly challenging case since the algorithm has to decide whether the new object has to inserted in front of the existing object or behind it.
In addition, if the new object needs to be inserted behind the existing object, then it also has to determine which part should be visible.
We note it requires scene parsing and understanding of 3D geometric to better infer how to seamlessly insert objects in videos, which will be our future work. 
It is also worth mentioning that our long-term goal is on video forensics (i.e., to detect fake or tampered videos) although we focus on inserting videos into videos in this work.

\section{User Study}

\begin{table}[t]
    \centering
    \caption{User study results on synthesized videos.
    Baseline 1 renders a video using a blending mode of the Adobe Premier CC Pro.
    Baseline 2 is based on a segmentation algorithm \cite{deeplabv3plus2018}.
    }
    \begin{tabular}{cccc}
    \toprule
     Method &
     Baseline 1 &
     Baseline 2 &
     Ours \\
     \midrule
     Avg. Score & 2.35 & 2.27 & \textbf{3.67} \\
     Preference & 17.3\% & 13.7\% & \textbf{70.0\%} \\
    \bottomrule
    \end{tabular}
    \label{tab:user}
\end{table}

\begin{figure}[t]
    \centering
    \includegraphics[width=1\linewidth]{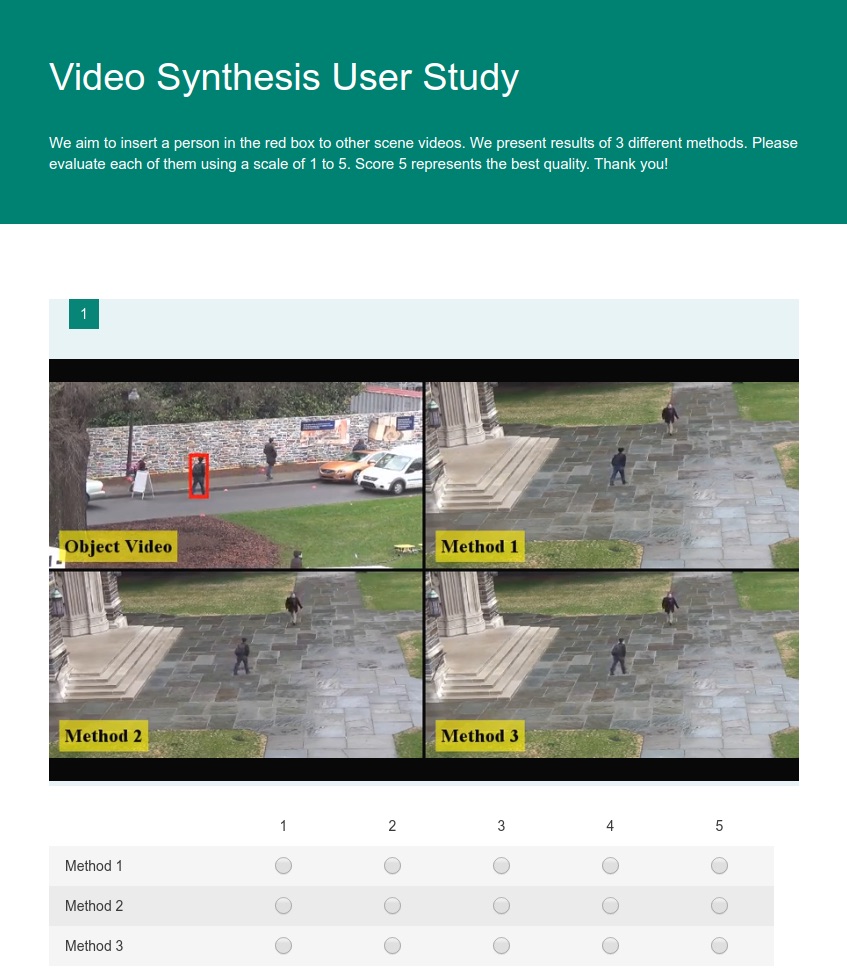}
    \caption{
    An example of our user study layout.
    In the middle, a user can play the video.
    }
\label{fig:user}
\end{figure}

We perform a human subject study to evaluate the realism of synthesized videos.
We conduct the experiments based on 22 test videos and 13 human workers.
Each video contains 300 frames (5 seconds) while descriptions of each algorithm are replaced by method 1, method 2, and method 3 as shown in Figure~\ref{fig:user}.
We ask workers to score each method from 1 to 5 (higher score for the better visual quality).
Therefore, each worker actually needs to assess 66 different results.
We provide two and three times slower videos with the original video to workers for more accurate evaluation.
Table~\ref{tab:user} shows the average score and percentage of cases that workers give the highest score to the method.
We find that for 70\% of the time the worker preferred our approach than baseline methods.
In addition, the proposed algorithm achieves significantly higher average scores.

\section{More Implementation Details}

\paragraph{Data preparation.}
The DukeMTMC dataset provides region of interest (ROI) to track pedestrians.
We use bounding boxes of pedestrians in the ROI as training and test data.
For $r_A$ and $r_B$, we pick a random location and a size around the ROI.
Then, we move $r_A$ by following a movement of a random pedestrian in video $A$.
We also scale the trajectory of the target object when it is inserted to video $A$ based on the height ratio between $r_B$ and $u_A$.
It is based on our assumption that the length of each step is approximately proportional to the height of a person.
For the TownCenter dataset, we use bounding boxes that are not cross the boundary of the image.
As the dataset does not provide ROI, we randomly sample a location to insert an object around the center of the image.

\begin{figure*}[t]
    \centering
    \includegraphics[width=.9\linewidth]{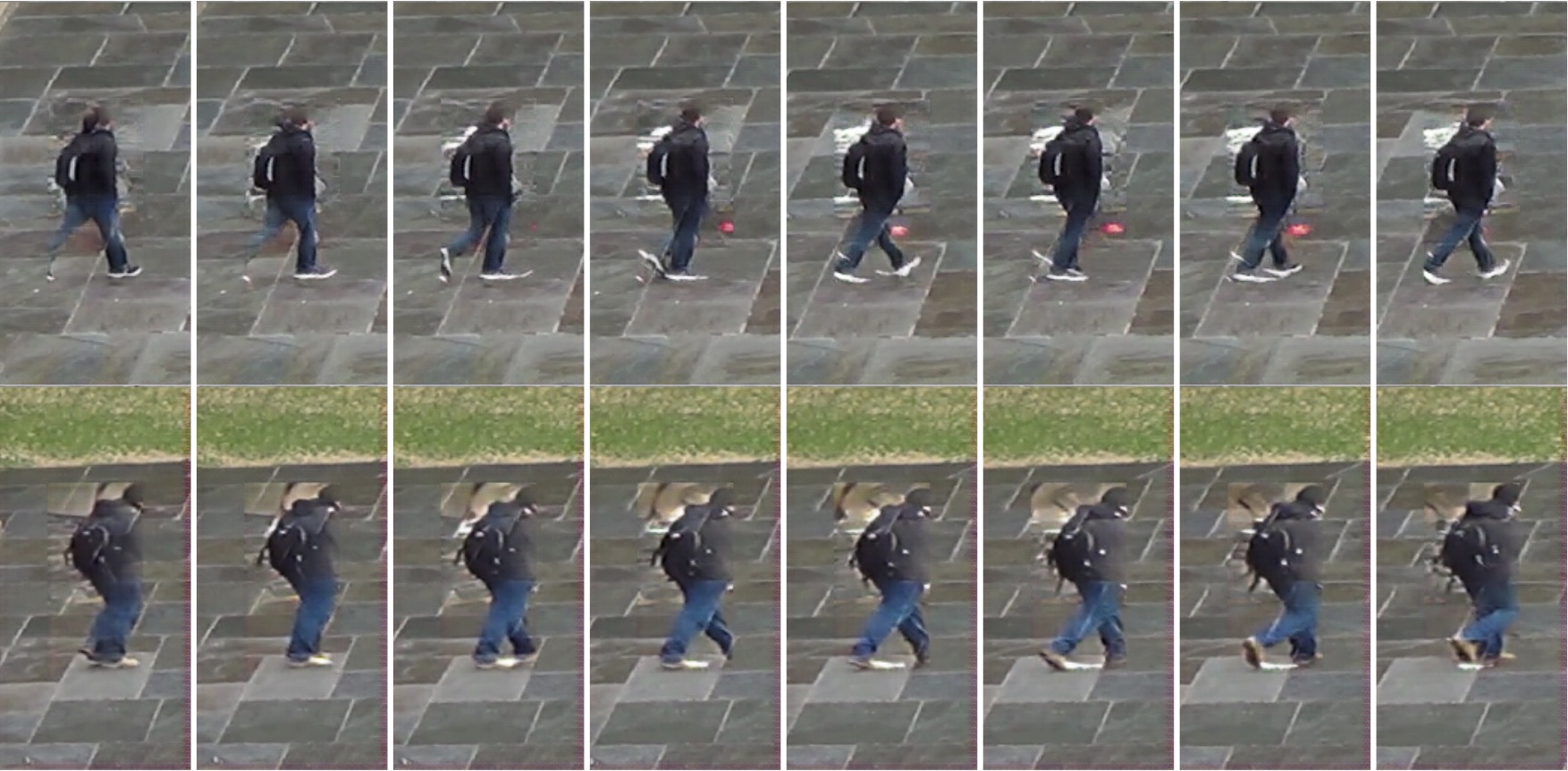}
    \caption{
    Results without noise injection to previous frames while rendering the current frame.
    It becomes easy for network to rely on previous frames and propagate wrong pixel values over time.
    }
\label{fig:noise}
\end{figure*}

\paragraph{Network training.}
While training, we use a parameter $\lambda$ to control the importance between the real and fake pairs.
It is multiplied with loss terms that are related to the fake pair.
Empirically we find that $\lambda=0.1$ makes the training process stable.
To make the training more stable, we inject noise to previous frames when generating the current frame as discussed in the paper.
Without the noise injection, the network blindly uses the information in the previous frame.
It may result in propagating wrong pixel values over time as shown in Figure~\ref{fig:noise}.
To address this issue, we add $0.01\times z$ at each pixel where $z$ is sampled from a normal Gaussian distribution.

\clearpage

{\small
\bibliographystyle{ieee}
\bibliography{videoinvideo}

\begin{thebibliography}{10}\itemsep=-1pt

\bibitem{bansal2018recycle}
A.~Bansal, S.~Ma, D.~Ramanan, and Y.~Sheikh.
\newblock Recycle-{GAN}: Unsupervised video retargeting.
\newblock In {\em European Conference on Computer Vision}, 2018.

\bibitem{benfold2011stable}
B.~Benfold and I.~Reid.
\newblock Stable multi-target tracking in real-time surveillance video.
\newblock In {\em IEEE Conference on Computer Vision and Pattern Recognition},
  2011.

\bibitem{chan2018everybody}
C.~Chan, S.~Ginosar, T.~Zhou, and A.~A. Efros.
\newblock Everybody dance now.
\newblock {\em arXiv preprint arXiv:1808.07371}, 2018.

\bibitem{deeplabv3plus2018}
L.-C. Chen, Y.~Zhu, G.~Papandreou, F.~Schroff, and H.~Adam.
\newblock Encoder-decoder with atrous separable convolution for semantic image
  segmentation.
\newblock In {\em European Conference on Computer Vision}, 2018.

\bibitem{chien2017detecting}
J.-T. Chien, C.-J. Chou, D.-J. Chen, and H.-T. Chen.
\newblock Detecting nonexistent pedestrians.
\newblock In {\em IEEE International Conference on Computer Vision}, 2017.

\bibitem{denton2017unsupervised}
E.~Denton and V.~Birodkar.
\newblock Unsupervised learning of disentangled representations from video.
\newblock In {\em Neural Information Processing Systems}, 2017.

\bibitem{finn2016unsupervised}
C.~Finn, I.~Goodfellow, and S.~Levine.
\newblock Unsupervised learning for physical interaction through video
  prediction.
\newblock In {\em Neural Information Processing Systems}, 2016.

\bibitem{gatys2016image}
L.~A. Gatys, A.~S. Ecker, and M.~Bethge.
\newblock Image style transfer using convolutional neural networks.
\newblock In {\em IEEE Conference on Computer Vision and Pattern Recognition},
  2016.

\bibitem{goodfellow2014generative}
I.~Goodfellow, J.~Pouget-Abadie, M.~Mirza, B.~Xu, D.~Warde-Farley, S.~Ozair,
  A.~Courville, and Y.~Bengio.
\newblock Generative adversarial nets.
\newblock In {\em Neural Information Processing Systems}, 2014.

\bibitem{hong2018learning}
S.~Hong, X.~Yan, T.~Huang, and H.~Lee.
\newblock Learning hierarchical semantic image manipulation through structured
  representations.
\newblock In {\em Neural Information Processing Systems}, 2018.

\bibitem{huang2017real}
H.~Huang, H.~Wang, W.~Luo, L.~Ma, W.~Jiang, X.~Zhu, Z.~Li, and W.~Liu.
\newblock Real-time neural style transfer for videos.
\newblock In {\em IEEE Conference on Computer Vision and Pattern Recognition},
  2017.

\bibitem{huang2018multimodal}
X.~Huang, M.-Y. Liu, S.~Belongie, and J.~Kautz.
\newblock Multimodal unsupervised image-to-image translation.
\newblock In {\em European Conference on Computer Vision}, 2018.

\bibitem{isola2017image}
P.~Isola, J.-Y. Zhu, T.~Zhou, and A.~A. Efros.
\newblock Image-to-image translation with conditional adversarial networks.
\newblock In {\em IEEE Conference on Computer Vision and Pattern Recognition},
  2017.

\bibitem{lee2018context}
D.~Lee, S.~Liu, J.~Gu, M.-Y. Liu, M.-H. Yang, and J.~Kautz.
\newblock Context-aware synthesis and placement of object instances.
\newblock In {\em Neural Information Processing Systems}, 2018.

\bibitem{liang2017dual}
X.~Liang, L.~Lee, W.~Dai, and E.~P. Xing.
\newblock Dual motion gan for future-flow embedded video prediction.
\newblock In {\em IEEE International Conference on Computer Vision}, 2017.

\bibitem{lin2018stgan}
C.-H. Lin, E.~Yumer, O.~Wang, E.~Shechtman, and S.~Lucey.
\newblock {ST}-{GAN}: {S}patial transformer generative adversarial networks for
  image compositing.
\newblock In {\em IEEE Conference on Computer Vision and Pattern Recognition},
  2018.

\bibitem{liu2016unsupervised}
M.-Y. Liu, T.~Breuel, and J.~Kautz.
\newblock Unsupervised image-to-image translation networks.
\newblock In {\em Neural Information Processing Systems}, 2017.

\bibitem{mathieu2015deep}
M.~Mathieu, C.~Couprie, and Y.~LeCun.
\newblock Deep multi-scale video prediction beyond mean square error.
\newblock In {\em International Conference on Learning Representations}, 2016.

\bibitem{ouyang2018pedestrian}
X.~Ouyang, Y.~Cheng, Y.~Jiang, C.-L. Li, and P.~Zhou.
\newblock Pedestrian-{S}ynthesis-{GAN}: {G}enerating pedestrian data in real
  scene and beyond.
\newblock {\em arXiv preprint arXiv:1804.02047}, 2018.

\bibitem{perez2003poisson}
P.~P{\'e}rez, M.~Gangnet, and A.~Blake.
\newblock Poisson image editing.
\newblock {\em ACM Transactions on Graphics}, 22(3):313--318, 2003.

\bibitem{radford2015unsupervised}
A.~Radford, L.~Metz, and S.~Chintala.
\newblock Unsupervised representation learning with deep convolutional
  generative adversarial networks.
\newblock {\em arXiv preprint arXiv:1511.06434}, 2015.

\bibitem{redmon2018yolov3}
J.~Redmon and A.~Farhadi.
\newblock Yolov3: {A}n incremental improvement.
\newblock {\em arXiv preprint arXiv:1804.02767}, 2018.

\bibitem{ristani2016MTMC}
E.~Ristani, F.~Solera, R.~Zou, R.~Cucchiara, and C.~Tomasi.
\newblock Performance measures and a data set for multi-target, multi-camera
  tracking.
\newblock In {\em European Conference on Computer Vision Workshops}, 2016.

\bibitem{ronneberger2015u}
O.~Ronneberger, P.~Fischer, and T.~Brox.
\newblock U-net: {C}onvolutional networks for biomedical image segmentation.
\newblock In {\em Proc. of the International Conference on Medical Image
  Computing and Computer-Assisted Intervention}, 2015.

\bibitem{ruder2018artistic}
M.~Ruder, A.~Dosovitskiy, and T.~Brox.
\newblock Artistic style transfer for videos and spherical images.
\newblock {\em International Journal of Computer Vision}, pages 1--21, 2018.

\bibitem{russakovsky2015imagenet}
O.~Russakovsky, J.~Deng, H.~Su, J.~Krause, S.~Satheesh, S.~Ma, Z.~Huang,
  A.~Karpathy, A.~Khosla, M.~Bernstein, et~al.
\newblock Imagenet large scale visual recognition challenge.
\newblock {\em International Journal of Computer Vision}, 115(3):211--252,
  2015.

\bibitem{simonyan15}
K.~Simonyan and A.~Zisserman.
\newblock Very deep convolutional networks for large-scale image recognition.
\newblock In {\em International Conference on Learning Representations}, 2015.

\bibitem{tesfaldet2018}
M.~Tesfaldet, M.~A. Brubaker, and K.~G. Derpanis.
\newblock Two-stream convolutional networks for dynamic texture synthesis.
\newblock In {\em IEEE Conference on Computer Vision and Pattern Recognition},
  2018.

\bibitem{villegas2017decomposing}
R.~Villegas, J.~Yang, S.~Hang, X.~Lin, and H.~Lee.
\newblock Decomposing motion and content for natural video sequence prediction.
\newblock In {\em International Conference on Learning Representations}, 2017.

\bibitem{villegas2017learning}
R.~Villegas, J.~Yang, Y.~Zou, S.~Hang, X.~Lin, and H.~Lee.
\newblock Learning to generate long-term future via hierarchical prediction.
\newblock In {\em International Conference on Machine Learning}, 2017.

\bibitem{vincent2008extracting}
P.~Vincent, H.~Larochelle, Y.~Bengio, and P.-A. Manzagol.
\newblock Extracting and composing robust features with denoising autoencoders.
\newblock In {\em International Conference on Machine Learning}, 2008.

\bibitem{walker2016uncertain}
J.~Walker, C.~Doersch, A.~Gupta, and M.~Hebert.
\newblock An uncertain future: {F}orecasting from static images using
  variational autoencoders.
\newblock In {\em European Conference on Computer Vision}, 2016.

\bibitem{wang2018video}
T.-C. Wang, M.-Y. Liu, J.-Y. Zhu, G.~Liu, A.~Tao, J.~Kautz, and B.~Catanzaro.
\newblock Video-to-video synthesis.
\newblock In {\em Neural Information Processing Systems}, 2018.

\bibitem{wen2015ua}
L.~Wen, D.~Du, Z.~Cai, Z.~Lei, M.-C. Chang, H.~Qi, J.~Lim, M.-H. Yang, and
  S.~Lyu.
\newblock {UA}-{DETRAC}: {A} new benchmark and protocol for multi-object
  detection and tracking.
\newblock {\em arXiv preprint arXiv:1511.04136}, 2015.

\bibitem{CycleGAN2017}
J.-Y. Zhu, T.~Park, P.~Isola, and A.~A. Efros.
\newblock Unpaired image-to-image translation using cycle-consistent
  adversarial networks.
\newblock In {\em IEEE International Conference on Computer Vision}, 2017.

\bibitem{zhu2017toward}
J.-Y. Zhu, R.~Zhang, D.~Pathak, T.~Darrell, A.~A. Efros, O.~Wang, and
  E.~Shechtman.
\newblock Toward multimodal image-to-image translation.
\newblock In {\em Neural Information Processing Systems}, 2017.

\end{thebibliography}
}

\end{document}